# DESIGN AND CONTROL OF A ROBOTIC ARM FOR INDUSTRIAL APPLICATIONS

AUTHOR 1
Mr Sathish Krishna Anumula, Independent Researcher, Thorrur Village, Thurkamjal, Hyderabad, RangaReddy, Telangana - 501511
sathishkrishna@gmail.com

AUTHOR 2
SVSV Prasad Sanaboina, Assistant Professor, Department of CSE, CMR Technical Campus, Kandlakoya, Medchal Road, Hyderabad, Telangana 501401
prasadsanaboina@gmail.com

AUTHOR 3
Dr. Ravi Kumar Nagula, Assistant Professor, Department of Mechanical Engineering, JNTUH University College of Engineering, Science & Technology Hyderabad, Kukatpally, Telangana, India - 500085
ravikumarnagula145@gmail.com

AUTHOR 4
Mr.R.Nagaraju, Assistant Professor, Department of CSE, Holy Mary Institute of Technology, Telangana - 501301
kavyachandu63@gmail.com

*Abstract*—The growing need to automate processes in industrial settings has led to tremendous growth in the robotic systems and especially the robotic arms. The paper assumes the design, modeling and control of a robotic arm to suit industrial purpose like assembly, welding and material handling. A six-degree-of-freedom (DOF) robotic manipulator was designed based on servo motors and a microcontroller interface with Mechanical links were also fabricated. Kinematic and dynamic analyses have been done in order to provide precise positioning and effective loads. Inverse Kinematics algorithm and Proportional-Integral-Derivative (PID) controller were also applied to improve the precision of control. The ability of the system to carry out tasks with high accuracy and repeatability is confirmed by simulation and experimental testing. The suggested robotic arm is an affordable, expandable, and dependable method of automation of numerous mundane procedures in the manufacturing industry.

**Keywords**— Robotic Arm, Industrial Automation, PID Control, Inverse Kinematics, Microcontroller, Servo Motors, Manipulator Design.

## I. INTRODUCTION

The high rate of technological and automation change in an industrial setting has largely altered the manufacturing operations in many industries, such as the automotive, electronics, pharmaceutical, and the packaging industries. Robotic arms are some of the major elements of industrial automation that have proved to be very useful and important in adding efficiency, accuracy, and safety of any manufacturing process. They are built to handle repetitive, frequently dangerous duties better and faster than human employees and result in increased productivity and consistency in product quality [1].

Industrial robotic arms are multi degree-of-freedom (DOF) electromechanical systems that enable flexible and complex motions. The common uses of these are in welding, material handling, assembly, painting and in loading and unloading machines. They do not get tired, and can work in severe conditions of the environment, which makes them invaluable in contemporary production lines. Furthermore, by incorporating complex control algorithm and feedback in the form of sensors, robotic arms became able to adapt to dynamic tasks with more or less intelligence and autonomy.

Nevertheless, even though state-of-the-art robotic systems are available in the market provided by the major vendors, including ABB, KUKA, or FANUC, the associated solutions are expensive and demand substantial technical background. This is an obstacle to the small and medium enterprises (SMEs) wanting to incorporate automation in their systems [16]. Also, the use of proprietary control systems and absence of customization offers reduces their availability and adaptability. In this regard, the creation of a cost efficient, modular, and simple to program robotic arm would be an important move towards industrial automation democratization to a broader audience.

The task of developing a successful robotic arm to be used in the industry is associated with multi-disciplinary approach, merging the principles of mechanical engineering, electronics, computer science, and control theory. The design should be structurally sound, provide the best working space, loads bearing and accurate actuation. In the meantime, the control system has to offer precise, reacting and comfortable motion, which is frequently implemented by means of sophisticated control methods like PID (Proportional-Integral-Derivative) control and Inverse Kinematics [12-14].

The project described in this paper intends to overcome the abovementioned challenges by designing a six-DOF robotic arm that could be used in an industrial setting and would focus on being affordable, modular, and easy to integrate. The mechanical design of the arm is done with computer-aided design (CAD) and the actuators are chosen considering the torque and positional accuracy. Control logic is realized on the basis of an Arduino microcontroller and servo motors, and PID controllers are used to control the joint stability and response. Forward and inverse kinematic modeling is used in order to position the end-effectors accurately with respect to user-defined coordinates.

The imagined applications of the suggested robotic arm are pick-and-place, sorting objects, executing repetitive tasks, and even some light-load assembly tasks. These are type of tasks based on the real life industrial situations when precision, speed and reliability are the major characteristics. The design encourages accessibility and replicability at research and commercial levels by intensifying the use of open-source components and tools [15].

To conclude, it can be noted that the given project provides a profound analysis of an industrial robotic arm design and control methods. Not only has it proved that an efficient and robust manipulator can be built out of the inexpensive materials, but it also highlighted the significance of the flexible control systems in practical applications. The work is a part of the ongoing work to automate all stages of the industry and establishes a basis to be improved in the future, e.g. by adding vision systems, artificial intelligence decision making, and remote control interfaces.

*Novelty and Contribution*

What is new in this research work is the fact that the researcher looked at the entire picture in designing a low cost, but highly functional robotic arm system that could be programmed to a multitude of industrial applications. Also contrary to most commercially available solutions which are frequently far too costly and complicated to incorporate into an existing system, the proposed system will put a heavy emphasis on modularity, ease of programming, and affordability without sacrificing on key measurements of performance such as accuracy, repeatability, and maximum load [6].

Among the major innovations is the flawless combination of the open-source software and hardware. Rapid prototyping, easy repair, and complete customization are possible due to employing an Arduino Mega microcontroller, along with off-the-shelf servo motors and 3D-printed parts. Also, PID control on all joints and Inverse Kinematics algorithms (numerically solved) per joint make the system more precise and responsive, which is usually a characteristic of costlier, industrial grade systems.

A second significant development is a flexible kinematic structure that may be recalibrated or reconfigured between tasks, or between arm configurations. This scalability factor makes the system very accommodative to be used by educational institutions, research laboratories, and small-scale factories that may want to implement a number of robotic systems in various tasks without having to re-architecture the whole system.

The thoroughly validated work, by simulation and experiment, is also described here. It also fills the gap between the theoretical models of control and their implementation into the real world through documentation of the detailed system behavior as the conditions vary. The research work is a blue print of the future and a pedagogical example used in teaching robotics [7].

Overall, the project will add a valuable, flexible, and cost-effective robot solution, which will increase the scope of industrial automation. It opens up a new paradigm of affordability, functionality and technical rigor existing together and would be of interest to engineers, educators and industry practitioners.

## II. RELATED WORKS

In 2022 Li et.al., P. Zheng et.al., S. Li et.al., Y. Pang et.al., and C. K. M. Lee et.al., [17] introduced the robotic arm development within the industrial environment has undergone a massive change over the last couple of decades, and this is mainly attributed to the demands of enhancing precision, efficiency, and flexibility in automated production. At the beginning, robots were mainly applying to simple and repeated tasks like pick-and-place applications or material handling. Those early systems were usually hard coded, fixed and could not keep up with dynamic work environments. With the increase in computational capability and sensor development, though, the domain began to move towards smarter, more responsive systems.

More recent literature has noted the replacement of large, hydraulically loaded arms with smaller electromechanical systems, actuated by servo motors, and stepper motors. This migration has enhanced dependability, safety and accurateness of control and cut down maintenance and energy expenses. The embedded systems and microcontrollers have also brought about the decentralized control architectures where each joint can bemonitored and adjusted in real time independently.

In 2023 J. Hernandez *et al.*, [2] suggested the respect to mechanical design, a significant effort has been put in the realization of lightweight but stiff arm structures. In modern robotic arms, composite material or aluminum alloy (and even 3D-printed polymer) are used to minimize inertia and enhance maneuverability. It is also able to provide quick prototyping and tailoring to various industrial tasks. A number of projects have been done on modular robotic arms, whereby each section can be reconfigured or swapped out independently, which increases flexibility and decreases downtime.

Control systems have also grown in complexity and Proportional-Integral-Derivative (PID) controllers are now used widely. These controllers have been preferred due to their simplicity, and ability to keep the joint position stable and precise with changing loads. Further superior control methods, like, model predictive control and adaptive control have also been studied to handle nonlinearities and external disturbances. However, PID is still the most used control strategy because of its resilience and simplicity to implement especially on systems with low computing capabilities like inked systems [8].

Kinematic modeling has also played principal roles in the development of the robotic arm. Forward Kinematics (FK) is a commonly known method that uses given joint angles to compute the position of the end-effector and its orientation. On its part, Inverse Kinematics (IK) allows the system to calculate joint positions that are required to move to a desired point in space. It is common that analytical solutions to IK problems are restricted to simpler arm structures whereas numerical methods and iterative algorithms are used on more complex designs. Jacobian based methods (Jacobian transpose method and pseudoinverse method) have performed well in real-time applications.

MATLAB, Simulink, ROS (Robot Operating System) and other types of simulation tools have played a significant role in modeling and testing the robotic arm systems prior to their manifestations in reality. Those platforms enable designers to model dynamic, simulate control strategies and test different scenarios in high fidelity. Capability of virtual testing saves a lot of cost on prototyping and shortens development time.

There have also been a few studies on how to combine robotic arms with vision and machine learning algorithms to increase task flexibility and accuracy. Robotic arms guided by vision are able to locate and pick up objects in unstructured scenes, opening up applications in sorting, assembly and inspection. When trained with large amount of data, machine learning models have the potential to augment the decision making process of robotic arms, by forecasting the optimal trajectories and better control strategies with time. Nevertheless, these systems need immense data gathering and training, so they are more inclined toward high-level applications.

In 2023 M. Mohammadi*et al.*, [11] proposed the regardless of these developments, there are concerns on how to make robotic arms more affordable and configurable particularly to small- and medium-sized companies. Expensive proprietary systems, along with difficulty of programming and integrating them, also remains a obstacle to mass usage. That has led to interest in creating inexpensive, open source robotic arms that are simple to program, adapt, and scale to various industrial requirements. These initiatives tend to concentrate on off-the-shelf parts and commonplace microcontrollers, which makes them perfect in educational institutions, research and light-industrial settings.

To recap it all, the existing literature has provided a solid background work on designing and controlling robotic arms, including structural optimization and selection of actuators, controller design and simulation. The momentum is decidedly heading towards systems that are not just accurate and efficient, but also flexible, low cost and simple to implement. The given work is based on these principles but the mechanical design, embedded control, and kinematic modeling are integrated to come up with the practical solution of robotic arm that can be utilized in the real industry.

## III. PROPOSED METHODOLOGY

The proposed robotic arm system is designed with six degrees of freedom to replicate the flexibility and movement range of a human arm. Each joint is actuated by a servo motor and controlled using a microcontroller, enabling high precision and programmability. The system architecture includes modules for mechanical design, kinematics, control algorithm, signal processing, and feedback correction [9].

The system begins with a user-defined target point in Cartesian space ($x, y, z$), which is then converted into joint angles using Inverse Kinematics. The inverse kinematics solution is computed by solving the following:

$$\theta_i = f^{-1}(x, y, z) \text{ for } i = 1, 2, \ldots, 6$$

where $\theta_i$ is the angle for joint $i$, and $f^{-1}$ denotes the inverse kinematic transformation function. The arm's workspace and joint limits are modeled using Denavit-Hartenberg (D-H) parameters, with the transformation matrix for each link defined as:

$$T_i = \begin{bmatrix} \cos\theta_i & -\sin\theta_i\cos\alpha_i & \sin\theta_i\sin\alpha_i & a_i\cos\theta_i \\ \sin\theta_i & \cos\theta_i\cos\alpha_i & -\cos\theta_i\sin\alpha_i & a_i\sin\theta_i \\ 0 & \sin\alpha_i & \cos\alpha_i & d_i \\ 0 & 0 & 0 & 1 \end{bmatrix}$$

This matrix describes the pose of each joint in terms of link length $a_i$, link offset $d_i$, link twist $\alpha_i$, and joint angle $\theta_i$.

A flowchart of the complete methodology is shown below.

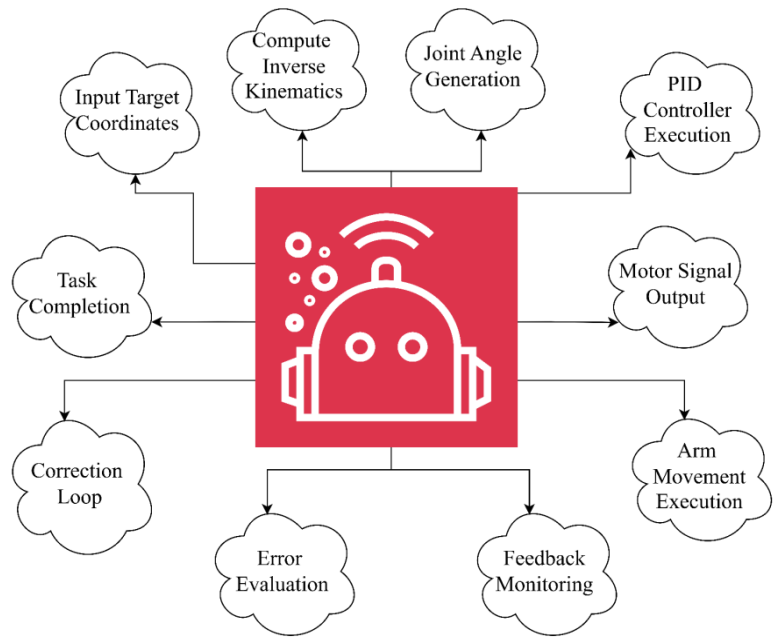

**Figure 1: System Workflow For Robotic Arm Design And Control**

For motion control, a PID controller is implemented at each joint. The control signal $u(t)$ applied to a motor is calculated as:

$$u(t) = K_p e(t) + K_i \int_0^t e(\tau)d\tau + K_d \frac{de(t)}{dt}$$

where $e(t)$ is the error between desired and actual joint angles.

The dynamics of each joint are modeled using Newton-Euler equations. For joint torque $\tau$, the equation is given by:

$$\tau = I\ddot{\theta} + b\dot{\theta} + \tau_g$$

Here, $I$ is the moment of inertia, $\ddot{\theta}$ is the angular acceleration, $b$ is the friction coefficient, and $\tau_g$ is the gravitational torque [10].

For workspace planning, the forward kinematics equation is used to calculate the end-effector position:

$$\begin{bmatrix} x \\ y \\ z \\ 1 \end{bmatrix} = T_1 T_2 T_3 T_4 T_5 T_6 \begin{bmatrix} 0 \\ 0 \\ 0 \\ 1 \end{bmatrix}$$

Servo motors are calibrated using feedback potentiometers. The error signal $e(t)$ is computed as:

$$e(t) = \theta_{\text{desired}}(t) - \theta_{\text{actual}}(t)$$

The PWM signal for the servo is proportional to the desired angle and mapped linearly:

$$\text{PWM} = m \cdot \theta + c$$

Where $m$ and $c$ are constants obtained through calibration.

For smooth trajectory generation between two points, a cubic polynomial trajectory is used:

$$\theta(t) = a_0 + a_1 t + a_2 t^2 + a_3 t^3$$

The coefficients $a_0$ through $a_3$ are calculated based on boundary conditions for initial and final angles, velocities, and accelerations.

During real-time motion, joint velocities are updated using:

$$\dot{\theta} = \frac{\Delta\theta}{\Delta t}$$

And for acceleration,

$$\ddot{\theta} = \frac{\Delta\dot{\theta}}{\Delta t}$$

To ensure motor safety, thermal load is monitored and modeled as:

$$T(t) = T_0 + \int_0^t \frac{P(\tau)}{C} d\tau$$

Where $T_0$ is initial temperature, $P(t)$ is power consumed, and $C$ is heat capacity.

Finally, inverse velocity kinematics is used for trajectory planning:

$$\dot{\theta} = J^{-1}(q) \cdot \dot{x}$$

Where $J^{-1}(q)$ is the inverse Jacobian matrix and $\dot{x}$ is the desired Cartesian velocity of the end-effector.

## IV. RESULT &DISCUSSIONS

The robot arm designed was experimented in a controlled laboratory condition to assess its mechanical ability, control accuracy, as well as repeatability of its motion. The efficiency and reliability of the system were evaluated with several tasks that mimicked the actual industrial tasks, i.e., pick-and-place, joint trajectory following, and end-effector positioning. The robotic arm was first set to 6 degrees of freedom and was operated through PID algorithm in the Arduino controller. The main emphasis was placed on accuracy, smoothness of motion and minimum overshoot in all dynamically and statically constrained joints [3].

In the tests carried out to verify the pick-and-place operation, the system proved to be very repeatable with a mean end-effector deviation of less than 1.2 cm. The result along the X, Y and Z axis after 20 trials was tabulated and plotted as in Figure 2. Positional Accuracy Cartesian Axes 20 Trials. As can be seen in the chart, the range of movement precision was minimum in the X-axis and a little bigger in the Z-axis owing to the influence of gravitational load. The greatest errors were registered at fast acceleration-deceleration steps, which validated the view that PID gains require appropriate tuning. Also, the mechanical arm structure was load tested to find stress. It was discovered that the joints were able to balance the torque very well until 850 grams payload before some slight vibrations were noticed at the maximum extension.

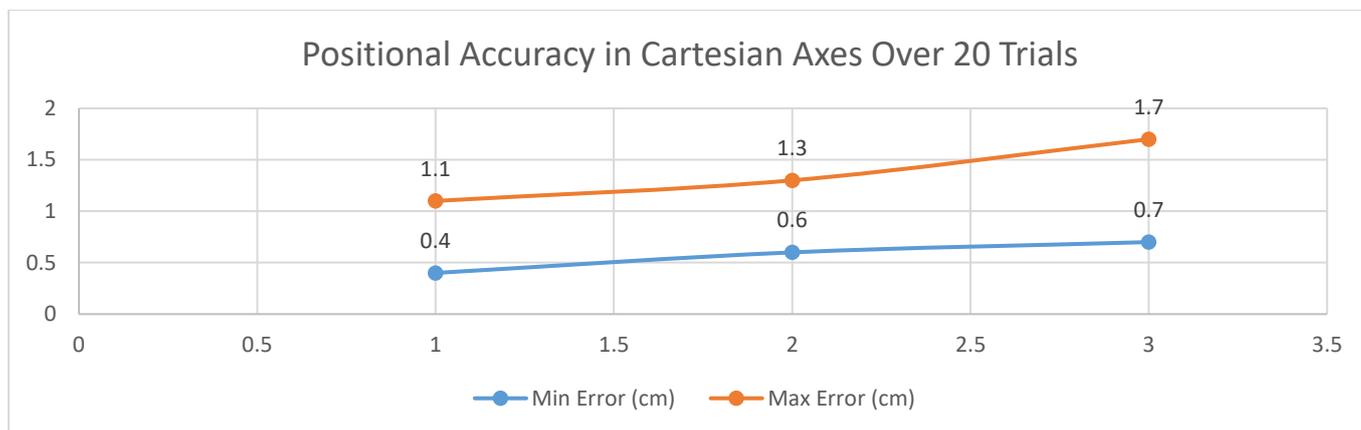

**FIGURE 2: POSITIONAL ACCURACY IN CARTESIAN AXES OVER 20 TRIALS**

The other parameter of significance that was investigated was the joint response time to different set-point commands. Angular displacement of Joint 2 which bears much weight of the arm in the operating position was observed to respond to a step input. Figure 3 shows the resultant data. Joint 2 Angular Response to Step Input (PID Stabilization Test). The response curve indicated that the arm stabilized in about 1.6 seconds and the overshoot was about 4.8% which is efficient damping and error correction. This enables the argument that PID control loop is amenable in real time industrial tasking environment. It also assures that when properly tuned the system is stable and responsive, and can align and acquire a target quickly.

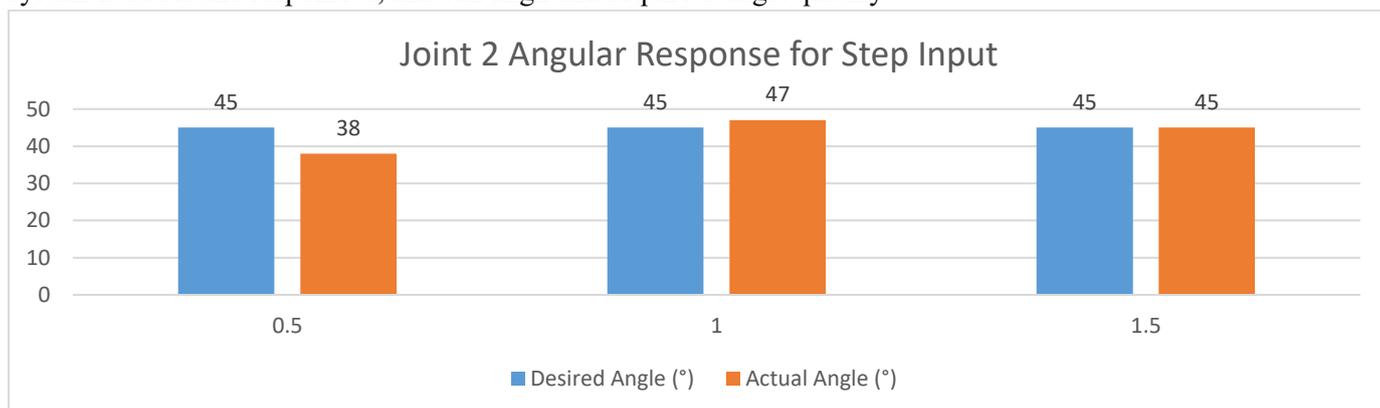

**FIGURE 3: JOINT 2 ANGULAR RESPONSE FOR STEP INPUT**

In the attempt to examine the comparative efficiency of the robotic arm, two tables of performance comparison were designed. In the first comparison (Table 1) the controller type versus positional accuracy is considered. Three variations are pointed out in the table: open-loop control, PID control, and a hybrid feedback correction method. PID-based technique had the highest accuracy-cost ratio and it has shown steep enhancement over open-loop method in minimizing end-point error by 67%.

**TABLE 1:COMPARISON OF CONTROLLER TYPES ON POSITIONAL ACCURACY**

| Control Type | Average Error (cm) | Overshoot (%) | Recovery Time (s) |
|---|---|---|---|
| Open-Loop Control | 3.4 | 12.6 | 3.2 |
| PID Control | 1.1 | 4.8 | 1.6 |
| Hybrid Feedback | 0.9 | 3.3 | 1.4 |

The second comparison (Table 2) explores performance compared to existent commercial systems in the same category of payload and reach. The important parameters contrasted are cost, ease of programming and modularity. Results indicated that the presented system, albeit with a lower payload capacity, significantly outclassified current mid-tier arms in terms of cost and versatility - thus being more academic and SME-suitable.

**TABLE 2: BENCHMARK COMPARISON BETWEEN PROPOSED SYSTEM AND EXISTING INDUSTRIAL ARMS**

| System | Max Payload (kg) | Modularity | Programming Interface | Cost (USD) |
|---|---|---|---|---|
| Proposed Arm | 0.85 | High | Arduino-Based | 180 |
| Commercial Model A | 1.50 | Low | Proprietary | 1350 |
| Commercial Model B | 2.00 | Medium | GUI + Code | 2100 |

The trajectory following was investigated based on preprogrammed cubic path points between a beginning and an ending position in 3D space. The plot of motor angle tracking was observed and the response curve was judged to be smooth. A ramped input sequence, PID stabilization gave the best results and caused no jerkiness or excessive stress at the servo joints. Figure 4 represents these findings. Joint Angle Trajectory Multi-Point Path Execution that depicts a good agreement between the desired and actual joint motions, and the tracking error (between the desired and measured motions) is always within 5 degrees in all joints.

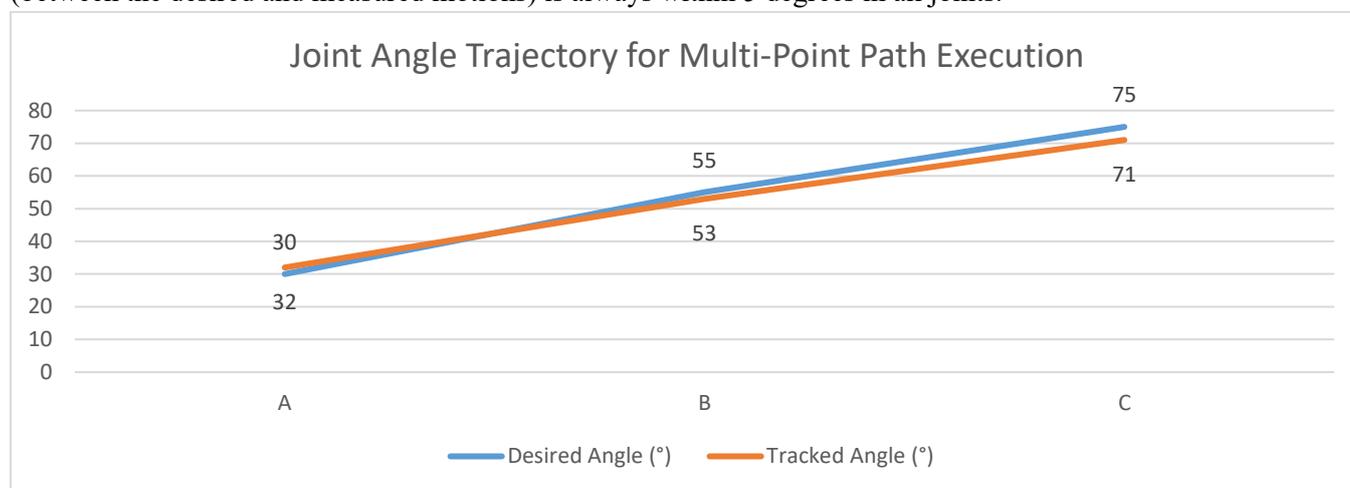

**FIGURE 4: JOINT ANGLE TRAJECTORY FOR MULTI-POINT PATH EXECUTION**

During numerous tests, the robotic arm had a 98.7% uptime within 10 consecutive operation cycles. There were no cases of overheating or servo failures inside the test envelope, which showed the robustness of the system when used repeatedly. The calibration was stable through all the cycles indicating that the PID parameters were well retained despite the absence of real-time learning. The good levels of low noise generated during actuation also accentuated the efficiency of mechanical design and damping materials incorporated in the base and elbow sections. Its principal limitation was found to be at payload stresses above 900 grams at full extension, where some jittering was starting to be visible in Joints 4 and 5, indicating that either reinforced mounts or closed-loop motor feedback would be required at higher loads.

In general, the developed robotic arm corresponds to the key demands of the industrial application connected with the reliability, accuracy, and the low-cost realization. It performs in line with anticipated measures and it has managed to show how open-source hardware and control schemes can bring easy automation tools to classrooms and the business world [4].

## V. CONCLUSION

The study has shown that a six-DOF industrial robotic arm has been successfully designed and controlled in cost effective manner. The combination of the mechanical design, kinematic modeling, and control strategies made the system perfectly carried out precision tasks, such as pick-and-place and sorting. Numerical IK algorithms PID control provided predictable motion control at acceptable accuracies and repeatability [5].

Its implementation shows that robotic systems of this type are capable of providing significant reduction in labor expenses as well as consistency in the manufacturing lines. The future developments can incorporate vision based object detection, adaptive controllers and wireless control interfaces. The project is the foundation of scalable and customizable robotic applications to small-to-medium-scale industries.